\documentclass[runningheads]{llncs}
\usepackage[T1]{fontenc}
\usepackage{graphicx}
\usepackage{algorithm}
\usepackage{algorithmic}
\usepackage{multirow}
\usepackage{subfig}

\usepackage{newfloat}
\usepackage{listings}

\usepackage{amsmath} 
\usepackage{booktabs}
\usepackage{pdflscape}

\usepackage{enumitem}
\begin{document}

\title{End-to-End Facial Expression Detection in Long Videos}
%
%
\author{Yini Fang\inst{1}\orcidID{0009-0008-9478-5545} \and
Alec F. Diallo\inst{2}\orcidID{0000-0002-0793-0492} \and
Frederic Jumelle\inst{3}\orcidID{0000-0002-9868-7167}
\and
Bertram Shi\inst{1}\orcidID{0000-0001-9167-7495}}
\authorrunning{Y. Fang et al.}
%
\institute{Hong Kong University of Science and Technology \and
University of Edinburgh \and
Ydentity Organization}
%
\maketitle              
\begin{abstract}
    Facial expression detection requires spotting when expressions occur and recognizing which emotional category they belong to. Despite their close relationships, existing approaches typically address these tasks separately, limiting performance and robustness in real-world settings. In this work, we propose FEDN, a Facial Expression Detection Network, which unifies spotting and recognition into a single detection task performed fully end-to-end. FEDN introduces two temporal attention modules, segment-level attention to capture fine-grained local dynamics and sliding window attention to capture the broader temporal context. Their output is combined in a multi-scale temporal feature pyramid, which enables spotting of expressions with varying duration. This unified framework enables joint optimization and shared representation learning across tasks. FEDN outperforms strong baselines in both spotting and detection on three public benchmarks, demonstrating the effectiveness of unifying spotting and recognition across multiple temporal scales. Additionally, we uncover a previously unreported discrepancy between expert-annotated and self-reported emotion labels, highlighting a key challenge in expression benchmarking and motivating the development of more nuanced annotation protocols.

\end{abstract}
\section{Introduction}
Facial expressions play a central role in conveying human emotions and social intents. Accurately interpreting expressions is crucial in fields, such as psychology, neuroscience, computer vision and human-computer interaction \cite{fang2023integrating}. Automatic facial expression analysis is typically divided into two tasks: \textit{spotting} (identifying the onset and offset of expressions) and \textit{recognition} (classifying the expression into an emotional category, e.g. happiness, anger, or fear). Despite their close relationships, these tasks have historically been studied in isolation \cite{thuseethan2023deep3dcann}\cite{liu2022clip}\cite{yuhong2021research}\cite{yin2023aware}\cite{liong2021shallow}. No prior work has proposed a joint model unifying both tasks in the context of facial expression analysis.

This separation stems from differences in annotation formats, modeling requirements, and evaluation protocols. The input to the spotting task is an entire video, with precise temporal labels. Models are optimized for fine-grained temporal localization. Recognition assumes pre-segmented clips and focuses on assigning a discrete emotional category to each clip. As a result, approaches to the two problems have evolved independently. 

However, spotting and recognition are clearly interdependent. Accurate recognition depends on reliable temporal boundaries. Conversely, semantic understanding of an expression can help to resolve ambiguous onsets and offsets. A unified model trained on both tasks can exploit a shared representation that captures these interdependencies.

While detection-based frameworks have been widely used for video event understanding and action recognition, applying them to facial expression analysis presents challenges: subtle and transient motions, sparse temporal signals, subjective emotional states, and high variability in expression duration. These characteristics make it non-trivial to transfer existing detection architectures.

In this work, we propose \textbf{FEDN}, an end-to-end \textbf{F}acial \textbf{E}xpression \textbf{D}etection \textbf{N}etwork, that simultaneously performs spotting and recognition in parallel. Previous work integrating spotting and classification \cite{gan2022needle,liong2023spot} considered the two tasks sequentially, with the spotter being trained first and then frozen before training the recognizer. Training both tasks simultaneously and in parallel allows the use of shared representations that leverage information from both tasks and enable the two tasks to co-adapt.

Importantly, we perform the first evaluation of facial expression detection and recognition as a unified task. We report detection metrics that simultaneously assess both temporal localization and emotion classification accuracy. In contrast, all prior works evaluated spotting and recognition separately. This enables a more realistic assessment of how well a model identifies and interprets expressions in continuous videos.

While 3D CNNs are commonly used for spotting~\cite{yu2023lgsnet}, they have high computational cost and limited ability to model long-term dependencies. To address this, FEDN introduces two lightweight temporal attention modules tailored for facial expression analysis: segment-level attention to fine-grained local changes and sliding window attention over a broader temporal context. These are combined with a multi-scale temporal feature pyramid to support expressions of varying durations. The resulting model is efficient, robust, and trained end-to-end.

In addition, our results reveal a systematic discrepancy between expert-annotated and self-reported emotion labels, a previously unidentified challenge to current expression benchmarks. This gap raises fundamental questions about the validity of ground-truth annotations and highlights the importance of labeling internal emotional states, not just observable facial cues.

Our contributions can be summarized as follows: 
\begin{enumerate}[nosep]
    \item We propose FEDN, the first model to unify facial expression spotting and recognition as parallel tasks.
    \item We perform the first evaluations of performance using detection metrics that reflect both localization and classification accuracy.
    \item We introduce segment-level and sliding window attention modules capturing short- and long-term facial dynamics.
    \item FEDN outperforms state-of-the-art spotting models on public benchmarks using appearance features only, without the need for computationally expensive motion cues, such as optical flow.
    \item We uncover new label source discrepancies and evaluate their impact on detection performance, motivating the development of more nuanced annotation protocols in future datasets.
\end{enumerate}

\vspace{-0.8em}
\section{Related Work}
\vspace{-0.5em}

Most prior work has examined either spotting or recognition in isolation. We first review these approaches before discussing efforts to integrate the two tasks.

\textbf{Spotting.}
Recent advances in deep learning have improved robustness by enabling data-driven modeling of temporal expression patterns \cite{liong2021shallow}, \cite{fang2023rmes}, \cite{yin2023aware}. LSSNet \cite{yu2021lssnet} marked a shift by reframing spotting as a detection task with IoU-based interval regression. LGSNet \cite{yu2023lgsnet} refined this with local suppression and global enhancement. Guo et al. \cite{guo2023micro} further improved temporal modeling by integrating Transformers into the feature pyramid. Despite these advances, most state-of-the-art spotters use I3D \cite{carreira2017quo} backbones pretrained on full-body action recognition. This results in a domain mismatch since facial expressions involve subtle, localized motion. In addition, 3D CNNs focus on fixed short-term temporal correlations, as they have high computational cost which becomes prohibitive as the window of temporal integration increases.

\textbf{Recognition.}
Facial expression recognition classifies emotions from predetermined segments of video frames containing expressions. Deep learning-based approaches use CNN-RNN hybrids, 3D CNNs, or multi-stream architectures to capture spatial and motion cues \cite{liu2022clip} \cite{yuhong2021research}. Some methods enhance recognition by selecting apex frames \cite{yin2023aware}, while others employ attention mechanisms to prioritize informative facial regions \cite{fang2023integrating}.

\textbf{Joint Spotting and Recognition.}
A few prior works have addressed both spotting and recognition, but only in a loosely coupled manner. Gan et al. \cite{gan2022needle} use localized peak detection for spotting and optical-flow-based descriptors for recognition. Liong et al. \cite{liong2023spot} propose a multi-stream model using shallow CNNs and handcrafted features like optical flow. In both cases, the two tasks are trained in sequence: the spotter is trained and frozen before the recognizer. This separation limits feature sharing and causes error propagation, as mis-localized intervals degrade recognition performance. In contrast, we reformulate expression analysis as a unified detection task, integrating spotting and recognition in a single end-to-end model. This enables joint feature learning via shared attention-based representations and removes the need for separate training stages. Unlike prior work, we evaluate detection holistically by capturing both when expressions occur and what they express, better aligning with real-world applications and demonstrating the advantages of integrated design.

\vspace{-0.5em}
\section{The Facial Expression Detection Network}
\vspace{-0.3em}


\begin{figure*}[t]
    \centering
    \includegraphics[width=0.78\textwidth]{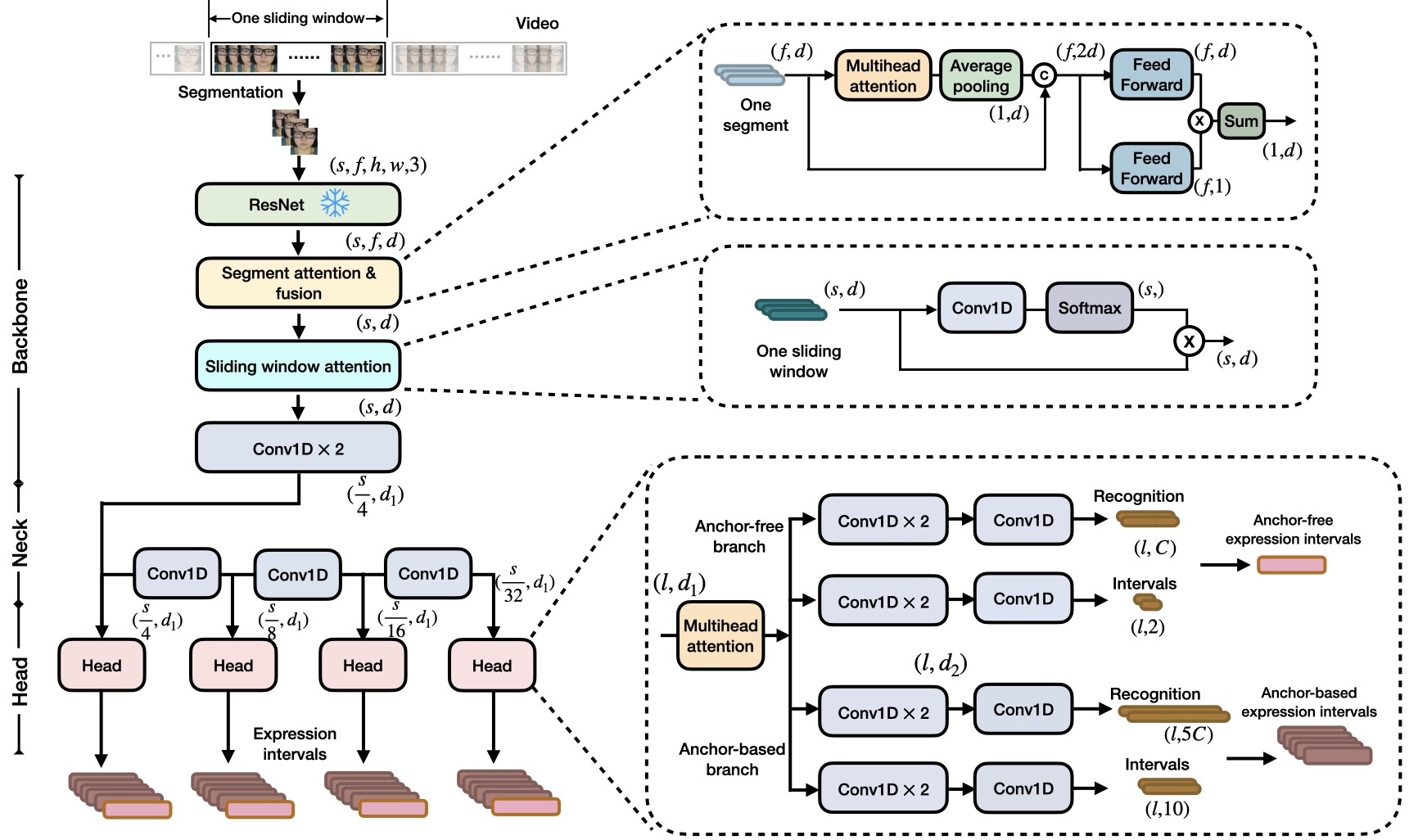}
    \caption{Overview of our facial expression detection network (FEDN). (($h, w, 3$): image dimensions, $s$: number of segments in sliding window, $f$: number of frames in segment, $l \in \{ s/4, s/8, s/16, s/32 \}$ depending upon the pyramid level,  ($d, d_1, d_2$): hidden layer dimensions, $C$: number of classes)
    }
    \label{main}
\end{figure*}


To process long video sequences efficiently, we adopt a sliding window mechanism. 
To capture short-term temporal dynamics, each window is divided into overlapping segments. Temporal reasoning is implemented with segment attention, which focuses on intra-segment motion, and sliding window attention, which operates across all the segments within the sliding window.

As illustrated in Figure~\ref{main}, our architecture consists of three modules: the backbone, neck, and head. The backbone first extracts spatial features from aligned face frames, then integrates information across short and long time scales using transformers. The space-time separable architecture results in not only lower computational complexity but also better performance, in comparison with more commonly used 3D CNN architectures. The neck constructs a multi-scale temporal feature pyramid to accommodate expressions of varying durations. The head outputs a set of expression intervals, confidence levels and emotion labels. We apply non-maximum suppression to remove overlapping boxes, producing the final outputs.


\textbf{Backbone.}
We use a pretrained facial feature extractor (e.g., ResNet18 trained on AffectNet~\cite{mollahosseini2017affectnet}) to encode each aligned face frame into a spatial feature vector. Features are taken from the penultimate layer to preserve rich representations without task-specific bias. This extractor is frozen during training.
Each frame with shape $(h, w, c)$ is encoded into a feature of dimension $d$, resulting in a sliding window representation of shape $(s, f, d)$, where $s$ is the number of segments and $f$ is the number of frames per segment.

To model short-term temporal structure, the feature sequence is passed through the segment attention module, which aggregates the $f$ frames within each segment. A global segment embedding is first computed using multi-head attention and average pooling. This vector is replicated and concatenated with each frame’s feature, forming a matrix of shape $(f, 2d)$. Each frame then passes through two parallel fully connected layers: one produces a scalar attention weight, and the other projects the features back to dimension $d$. The segment embedding is formed by a weighted sum across frames, yielding an output of shape $(s, d)$.
Next, the sliding window attention module enhances long-range context by applying attention across the segment sequence. Attention scores are computed via a 1D convolution followed by Softmax, and applied via element-wise multiplication to the sequence.
Finally, the refined segment features are passed through two 1D CNN blocks (stride 2), producing a downsampled sequence of length $\frac{s}{4}$.

\textbf{Neck.}
Facial expressions vary widely in duration, making it crucial to detect intervals at multiple temporal scales. To address this, we design the neck to construct a temporal feature pyramid with four resolution levels. We apply a series of 1D convolutional layers with stride 2 to the backbone output, giving feature sequences of length $\in \frac{s}{4}, \frac{s}{8}, \frac{s}{16}, \frac{s}{32}$.

\textbf{Head.}
The head module predicts six sets of start and end positions of expression intervals and associate confidence intervals and expression labels at each temporal location and each temporal scale, denoted by  $\hat{B}=(\hat{y}_s, \hat{y}_e, \hat{c}, \hat{e})$,
where $\hat{y}_s$ and $\hat{y}_e$ denote the relative start and end locations within the sliding window, $\hat{c}$ is the confidence score, and $\hat{e}$ is the predicted emotion category. Scales are processed independently. 

To capture localized temporal dependencies specific to each resolution, each branch begins with multi-head attention, focusing on context around each temporal position. This is followed by 1D CNN layers that reduce dimensionality to $d_2$. To improve detection flexibility and localization precision, we use both anchor-free and anchor-based predictions: One interval is predicted anchor-free, directly regressing start and end locations relative to the temporal location. Five intervals are predicted anchor-based, using predefined anchor templates based on interval statistics from prior datasets, following the design of LGSNet \cite{yu2023lgsnet}.




\textbf{Post-processing. }
Applying the model across all sliding windows in a video results in a large number of overlapping expression interval predictions. To reduce redundancy and ensure coherent output, we apply non-maximum suppression (NMS) at the video level, which retains only the highest-confidence  intervals, consolidates predictions from multiple windows into a final, non-redundant set of detected expressions.

\textbf{Learning Objectives. }
Each predicted expression interval is assigned a binary label during training. A positive interval matches a ground-truth expression within a tolerance threshold, and inherits its corresponding label and temporal boundaries. A negative interval corresponds to a prediction that overlaps no ground-truth expression (e.g., neutral or background segments). The learning objectives are defined as follows:

\begin{enumerate}[nosep]
    \item \textit{Interval loss ($\mathcal{L}_b$):} The 1D Distance IoU (DIoU) loss \cite{zheng2020distance} measures the alignment between predicted and ground truths for positive intervals. For negative intervals, this loss is set to zero.

    \item \textit{Recognition loss ($\mathcal{L}_c$):}We apply binary cross-entropy (BCE) loss to the classification scores, treating the highest scoring category as the predicted emotion. This formulation accommodates the multi-label nature of detection, where multiple candidate intervals may co-occur and be scored in parallel.

\end{enumerate}

The total loss across all $M$ predicted intervals is computed as a weighted sum of the individual losses:
$\mathcal{L} = \frac{1}{M} \sum^{M}_{i=1} ( \mathcal{L}_b^i + \beta \mathcal{L}_c^i )$,
where $\beta$ is an empirically chosen hyperparameter balancing localization and recognition.

\section{Experiment}

\textbf{Datasets.}
We evaluate FEDN on three datasets.
$\text{CAS(ME)}^2$ includes 98 videos and 300 expressions with both expert and self-reported labels\cite{qu2017cas}. 
$\text{CAS(ME)}^3$ contains 1300 videos and 3346 expressions with expert annotations\cite{li2022cas}. 
SAMMLong provides 159 high-FPS samples with spotting labels only\cite{yap2020samm}. We report detection results on $\text{CAS(ME)}^2$ and $\text{CAS(ME)}^3$, and spotting results on SAMMLong. Table \ref{emotion} summarizes the label characteristics for the $\text{CAS(ME)}^2$ dataset.

\begin{table*}[t]
    \centering
    \small
    \caption{Emotion distributions in CAS(ME)$^2$
    show a discrepancy between
    annotated and self-reported labels.}
    \setlength{\tabcolsep}{4pt}

    \begin{tabular}{l l l l l l l l l l l}
        \toprule

        \textbf{CAS(ME)$^2$ Annotated}
         & positive 116
         & surprise 16
         & negative 105
         & others 63     \\
         & Total 300     \\

        \midrule

        \textbf{CAS(ME)$^2$ Self-reported}
         & happiness 132
         & surprise 26
         & disgust 58
         & anger 47      \\
         & fear 17
         & sadness 8
         & helpless 4
         & confused 3    \\
         & pain 3
         & sympathy 2
         & Total 300     \\



        \bottomrule
        \label{emotion}
    \end{tabular}
\end{table*}

\begin{table*}[t]
    \centering
    \small
    \caption{Performance comparison (\textbf{\textit{S}}=spotting, \textbf{\textit{R}}=recognition, \textbf{\textit{D}}=detection). `-' indicates not applicable. (spo)/(rec) indicate models trained only with spotting/recognition labels.}
    \begin{tabular}{cc|cccccccc|cccc}
        \midrule[1.2pt]
        \multicolumn{1}{c|}{\multirow{4}{*}{Spotter}} & \multirow{4}{*}{Recognizer} & \multicolumn{8}{c|}{\textbf{CAS(ME)$^2$}}                                                               & \multicolumn{4}{c}{\textbf{CAS(ME)$^3$}}                                                                                                                                                                                                                                                                                                                                                                                                                                                                                                                                                                                                                                                                                                                        \\ \cline{3-14}
        \multicolumn{1}{c|}{}                         &                             & \multicolumn{4}{c|}{Annotated}                                                                          & \multicolumn{4}{c|}{Self-reported}                                                                       & \multicolumn{4}{c}{Annotated}                                                                                                                                                                                                                                                                                                                                                                                                                                                                                                                                                                                                                        \\ \cline{3-14}
        \multicolumn{1}{c|}{}                         &                             & \multicolumn{2}{c|}{\begin{tabular}[c]{@{}c@{}}F1\\ ($10^{-2}$)\end{tabular}} & \multicolumn{2}{c|}{\begin{tabular}[c]{@{}c@{}}mAP\\ ($10^{-3}$)\end{tabular}} & \multicolumn{2}{c|}{\begin{tabular}[c]{@{}c@{}}F1\\ ($10^{-2}$)\end{tabular}} & \multicolumn{2}{c|}{\begin{tabular}[c]{@{}c@{}}mAP\\ ($10^{-3}$)\end{tabular}} & \multicolumn{2}{c|}{\begin{tabular}[c]{@{}c@{}}F1\\ ($10^{-2}$)\end{tabular}} & \multicolumn{2}{c}{\begin{tabular}[c]{@{}c@{}}mAP\\ ($10^{-3}$)\end{tabular}}                                                                                                                                                                                                             \\ \cline{3-14}
        \multicolumn{1}{c|}{}                         &                             & \textbf{\textit{S}}                                                                                     & \multicolumn{1}{c|}{\textbf{\textit{R}}}                                                                 & \textbf{\textit{S}}                                                                                     & \multicolumn{1}{c|}{\textbf{\textit{D}}}                                                                 & \textbf{\textit{S}}                                                                                      & \multicolumn{1}{c|}{\textbf{\textit{R}}}                                                                 & \textbf{\textit{S}}                & \multicolumn{1}{c|}{\textbf{\textit{D}}} & \textbf{\textit{S}}                & \multicolumn{1}{c|}{\textbf{\textit{R}}} & \textbf{\textit{S}} & \textbf{\textit{D}} \\ \midrule[1.2pt]
        \multicolumn{2}{c|}{STR}                      & 19.5                        & \multicolumn{1}{c|}{58.9}                                                                               & 11.4                                                                                                     & \multicolumn{1}{c|}{13.4}                                                                               & 22.0                                                                                                     & \multicolumn{1}{c|}{48.7}                                                                                & 17.4                                                                                                     & \multicolumn{1}{c|}{1.4}           & 12.3                                     & \multicolumn{1}{c|}{23.3}          & 5.3                                      & 2.0                                       \\ \cline{1-2}
        \multicolumn{1}{c|}{FEDN (spo)}               & CEFLNet                     & 50.7                                                                                                    & \multicolumn{1}{c|}{71.2}                                                                                & 77.0                                                                                                    & \multicolumn{1}{c|}{42.1}                                                                                & 50.7                                                                                                     & \multicolumn{1}{c|}{67.2}                                                                                & 77.0                               & \multicolumn{1}{c|}{10.9}                & 34.5                               & \multicolumn{1}{c|}{39.4}                & 52.5                & 4.2                 \\ \cline{1-2}
        \multicolumn{1}{c|}{FEDN (spo)}               & STR                         & 50.7                                                                                                    & \multicolumn{1}{c|}{73.6}                                                                                & 77.0                                                                                                    & \multicolumn{1}{c|}{42.5}                                                                                & 50.7                                                                                                     & \multicolumn{1}{c|}{67.2}                                                                                & 77.0                               & \multicolumn{1}{c|}{11.1}                & 34.5                               & \multicolumn{1}{c|}{25.6}                & 52.5                & 2.7                 \\ \cline{1-2}
        \multicolumn{1}{c|}{FEDN}                     & CEFLNet                     & 51.1                                                                                                    & \multicolumn{1}{c|}{72.9}                                                                                & 106                                                                                                     & \multicolumn{1}{c|}{45.5}                                                                                & 52.4                                                                                                     & \multicolumn{1}{c|}{70.5}                                                                                & 102                                & \multicolumn{1}{c|}{17.6}                & 35.0                               & \multicolumn{1}{c|}{38.3}                & 54.3                & 5.3                 \\ \cline{1-2}
        \multicolumn{1}{c|}{FEDN}                     & STR                         & 51.1                                                                                                    & \multicolumn{1}{c|}{72.8}                                                                                & 106                                                                                                     & \multicolumn{1}{c|}{49.7}                                                                                & 52.4                                                                                                     & \multicolumn{1}{c|}{74.5}                                                                                & 102                                & \multicolumn{1}{c|}{14.8}                & 35.0                               & \multicolumn{1}{c|}{25.9}                & 54.3                & 3.1                 \\ \midrule[0.8pt]
        \multicolumn{1}{c|}{FEDN (spo)}               & FEDN (rec)                  & 50.7                                                                                                    & \multicolumn{1}{c|}{71.1}                                                                                & 77.0                                                                                                    & \multicolumn{1}{c|}{43.8}                                                                                & 50.7                                                                                                     & \multicolumn{1}{c|}{66.8}                                                                                & 77.0                               & \multicolumn{1}{c|}{13.5}                & 34.5                               & \multicolumn{1}{c|}{-}                   & 52.5                & -                   \\ \cline{1-2}
        \multicolumn{2}{c|}{\textbf{FEDN}}            & \textbf{51.1}               & \multicolumn{1}{c|}{\textbf{75.2}}                                                                      & \textbf{106}                                                                                             & \multicolumn{1}{c|}{\textbf{51.6}}                                                                      & \textbf{52.4}                                                                                            & \multicolumn{1}{c|}{\textbf{74.6}}                                                                       & \textbf{102}                                                                                             & \multicolumn{1}{c|}{\textbf{18.4}} & \textbf{35.0}                            & \multicolumn{1}{c|}{\textbf{40.9}} & \textbf{54.3}                            & \textbf{5.4}                              \\ \midrule[1.2pt]
        \label{performance}
    \end{tabular}
\end{table*}


\begin{table*}[t!]
    \centering
    \small
    \caption{F1 scores for spotting task on CAS(ME)$^2$ and SAMMLV}
    \label{spot_res}

    \begin{tabular}{lccccccc}
        \toprule
         & LSSNet         & MTSN & Tan et al. & AUW-GCN & LGSNet & SpotFormer & \textbf{FEDN} \\
        \midrule
        CAS(ME)$^2$
         & 0.380
         & 0.410
         & 0.424
         & 0.424
         & 0.464
         & 0.506
         & \textbf{0.524}                                                                     \\

        SAMMLV
         & 0.281
         & 0.346
         & 0.423
         & 0.429
         & 0.388
         & 0.445
         & \textbf{0.447}                                                                     \\

        \bottomrule
        \label{spot_res}
    \end{tabular}
\end{table*}

\textbf{Implementation Details.}
Each face is cropped to $224 \times 224$ pixels. The sliding window contains $s{=}64$ segments, each with $f{=}8$ frames and 6-frame overlap. Feature dimensions are $d{=}512$, $d_1{=}512$, and $d_2{=}256$. We use loss weight $\beta{=}2$, and evaluate with LOSO cross-validation. Training uses Adam optimizer (lr{=}0.0001, weight decay{=}0.0001).

\textbf{Metrics.}
1) \textbf{F1 score} \cite{li2019spotting} For recognition, F1 scores are computed only on correctly spotted intervals. As a result, F1 scores depend on the spotter’s accuracy and are not directly comparable across models using different spotters. 2) \textbf{Mean Average Precision (mAP)} We compute the AP@[.5:.95] metric following the MS COCO object detection challenge \cite{lin2014microsoft}. Unlike F1 which uses a fixed IoU threshold, mAP evaluates performance under varying overlap tolerances, offering a more comprehensive performance assessment.

\vspace{-0.5em}
\section{Results}

\subsection{Performance Comparison}

We categorize the baselines into three groups:
\begin{itemize}[nosep]
    \item \textit{Spotting Models}: LSSNet~\cite{yu2021lssnet}, MTSN~\cite{liong2022mtsn}, Tan et al.\cite{tan2023unbiased}, AUW-GCN\cite{yin2023aware}, and SpotFormer~\cite{deng2024spotformer}. These methods do not address recognition.
    \item \textit{Detection Models:}  STR~\cite{liong2023spot} is one of the only two prior works that generate both temporal intervals and expression labels for long videos.
    \item \textit{Cascaded Spotter + Recognizer}: We build two-stage pipelines using FEDN’s spotter (trained with or without recognition labels), combined with other state of the art recognizers: CEFLNet~\cite{liu2022clip}, a CNN-based sequential model that processes the frames from onset to offset, or STR’s recognizer. This setup illustrates the impact of joint supervision.

\end{itemize}

Table~\ref{performance} summarizes our model’s performance across detection and recognition tasks. FEDN achieves state-of-the-art results on all datasets and metrics. On CAS(ME)$^2$ (annotated labels), it improves the detection mAP from 13.4 to 51.6 compared to STR, a 3.9-fold increase, while also achieving the highest spotting and recognition F1 scores. Similar gains are observed on CAS(ME)$^3$, demonstrating strong generalization.

\textbf{Spotting-Only Evaluation.}
Table~\ref{spot_res} shows that FEDN, even when trained without recognition supervision, achieves the best spotting performance on the CAS(ME)$^2$ and SAMMLong datasets, with 3.6\% larger F1 than SpotFormer. Unlike prior models that rely on I3D features pretrained for action recognition, FEDN uses a lightweight space-time separable facial feature backbone with task-aware attention, improving precision on subtle facial movements.

\textbf{Comparison with Detection Models.}
STR underperforms across all metrics due to two main limitations. First, its spotting module uses optical flow between onset and apex frames only, missing key temporal dynamics. Second, feature extraction in STR’s recognizer is not optimized jointly with its spotter, limiting its ability to align with localization errors. In contrast, FEDN attends to all frames in a sliding window and shares features across tasks, enabling more accurate and coherent detection without the computational overhead of optical flow.

\begin{figure*}[t]
    \centering
    \subfloat[Spotting AP]{\label{a}\includegraphics[width=.3\linewidth]{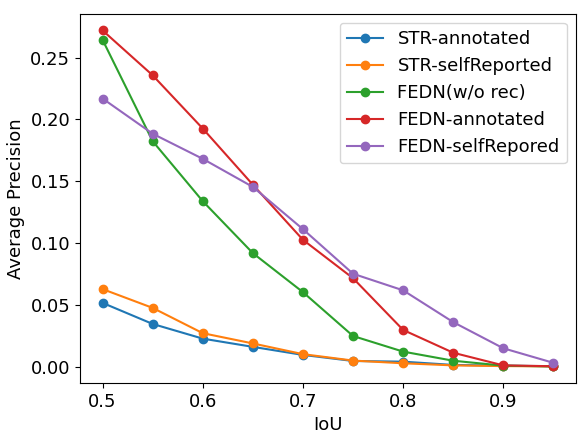}}
    \subfloat[Detection AP (annotated)]{\label{b}\includegraphics[width=.3\linewidth]{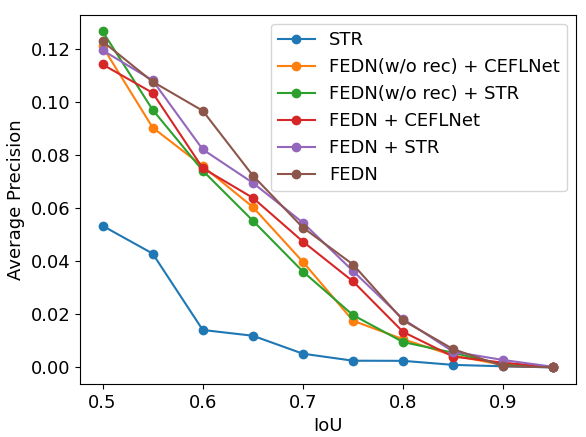}}
    \subfloat[Detection AP (self-reported)]{\label{c}\includegraphics[width=.3\linewidth]{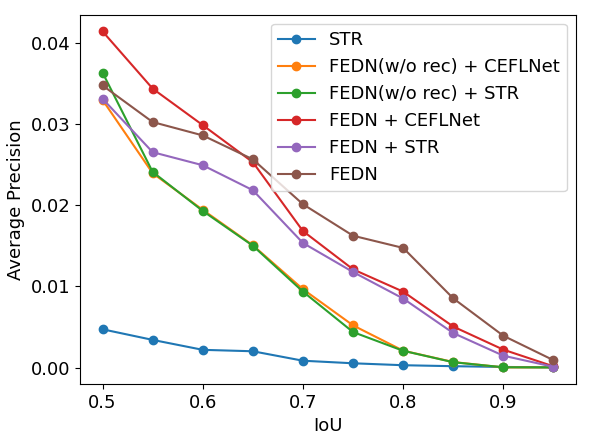}}
    \caption{AP vs. IoU curves for spotting and detection on CAS(ME)$^2$.
    }
    \label{aps}

\end{figure*}

\textbf{Cascaded vs. Unified Models.}
Both cascaded pipelines (spotter + recognizer) yield lower detection mAP than our unified model. While the recognizers perform well when given ground-truth intervals, they degrade when used on predicted segments. This highlights a key weakness of cascaded systems: they cannot adapt to noisy or imperfect localizations. In contrast, FEDN is trained end-to-end, allowing shared representations to evolve jointly across tasks, resulting in more robust detection performance.

\textbf{Effect of Joint Supervision.}
Training FEDN with recognition labels improves both spotting and detection accuracy. As shown in Table~\ref{performance}, removing recognition supervision reduces the spotting F1 scores across all datasets. This supports our hypothesis that emotion labels help resolve ambiguous cases and guide the attention mechanisms to focus on semantically meaningful frames.

\subsection{Detailed Comparison and Analysis}
\label{detailed_comparison}

\textbf{Spotting AP vs. IoU.}
Figure~\ref{aps}(a) presents the AP of all models on CAS(ME)$^2$ across different IoU thresholds. Notably, FEDN trained with self-reported labels outperforms the version trained on annotated labels at higher IoU thresholds. We hypothesize that self-reported labels more accurately reflect the subject’s internal emotional state, reducing ambiguity in interval boundaries. Annotated labels are assigned by third-party coders viewing the video without knowledge of the subject’s feelings. Misinterpretation introduced noise into supervision. This discrepancy is evident in Table~\ref{emotion}, where the distribution of categories like “positive” and “surprise” varies substantially between the two label types. Interestingly, at lower IoU thresholds, the trend reverses. FEDN trained with annotated labels achieves better AP. We attribute this to the tendency of third-party annotators to overextend interval boundaries, increasing overlap and boosting low-IoU tolerance performance.




\textbf{Detection AP vs. IoU.}
Figures~\ref{aps}(b) and \ref{aps}(c) show detection AP under varying IoU thresholds for annotated and self-reported labels, respectively. Across both settings, FEDN outperforms all compared baselines over nearly the entire IoU range. This confirms the benefits of jointly optimizing spotting and recognition: shared temporal features improve localization precision. However, when the spotter is inaccurate (especially at IoU$<$0.65), the detection AP of a cascaded pipeline (FEDN spotter + CEFLNet recognizer) slightly exceeds that of the joint model. This suggests that joint training may propagate spotting errors into recognition, reducing robustness under low-quality localization. Nonetheless, FEDN consistently performs best at stricter IoU thresholds, demonstrating its advantage in precise expression detection.

\begin{figure*}[t]
    \centering
    \includegraphics[width=0.8\linewidth]{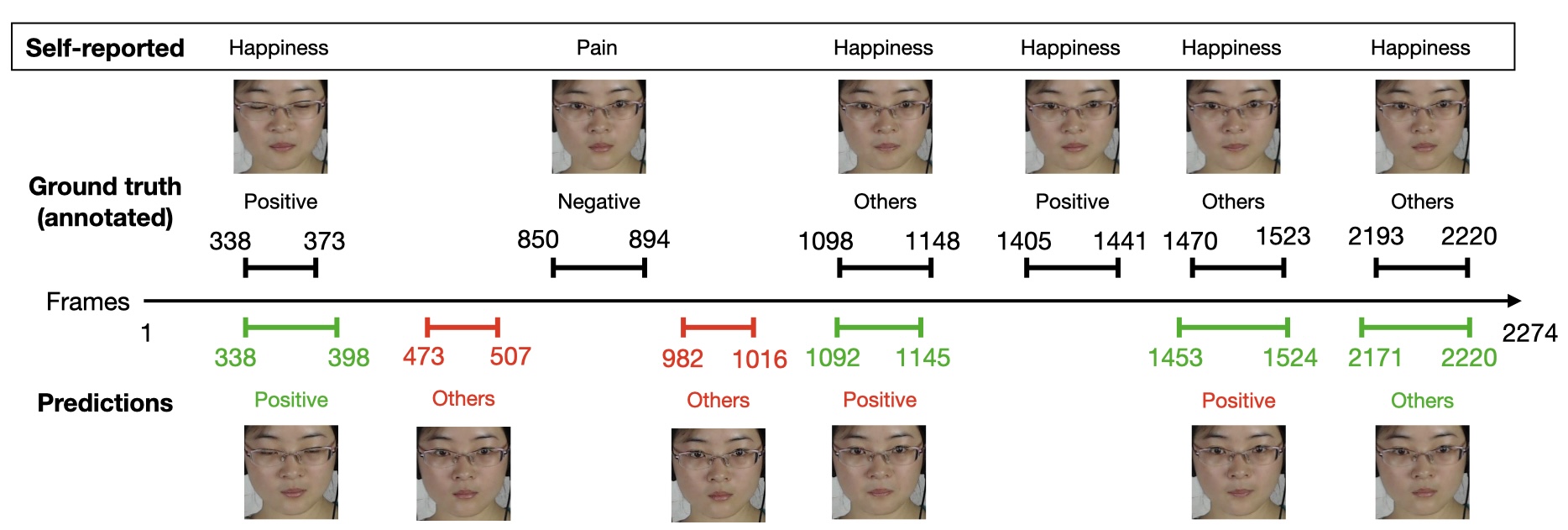}
    \caption{Timeline-based comparison of predicted expression intervals and ground truth annotations in CAS(ME)$^2$, after training with the expert-annotated labels. The top row shows ground truth intervals from both expert and self-reported labels. The bottom row shows model predictions. Correct predictions are in green; errors are in red. The video spans 2,274 frames (75.8s) with six annotated expressions. In the last four cases, the subject consistently self-reports “happiness,” aligning with the model’s predictions, while the expert labels them as “others”.}
    \label{vis}
\end{figure*}

\textbf{Qualitative Visualization.}
Figure~\ref{vis} illustrates a timeline comparison of FEDN's predicted expression intervals against expert-annotated and self-reported ground truth labels in CAS(ME)$^2$. The model correctly detects four intervals, with two false positives and two false negatives. Notably, in the last four intervals, the subject consistently self-reports “happiness”, which aligns with the model’s predictions, while the expert labels them as “others”. Self-reports reflect the subject’s internal emotional state, whereas expert labels rely on third person interpretation of facial cues. These labels can diverge, especially when expressions are subtle. A more systematic analysis (e.g., inter-annotator agreement metrics) could strengthen the claim. This open question is an opportunity for future dataset design and benchmarking efforts.

\subsection{Ablation Study}

To assess our feature extraction strategy, we conducted an ablation study (Table~\ref{ablation}). We compare FEDN to (1) an I3D backbone pretrained on action recognition, and (2) FEDN variants with different attention modules removed. FEDN outperforms I3D in detection mAP (51.6 vs. 35) with fewer parameters (24.98M vs. 30.52M), highlighting the efficiency of our lightweight design tailored for facial expression analysis. Removing segment attention (\textit{seg. att.}) or sliding-window attention (\textit{sw att.}) reduces mAP to 47 and 43, respectively, while removing both drops it to 38. These modules contribute a 35.8\% gain over the no-attention baseline. We also tested unfreezing the last two ResNet18 blocks. Without attention, this improves mAP from 38 to 46, but still underperforms full FEDN (51.6) while using more parameters. Unfreezing with full FEDN gives a marginal mAP boost to 54.4 but increases parameters by 33\% (to 33.38M). This confirms attention is more effective and efficient than backbone fine-tuning alone.

\begin{table}[t!]
    \centering
    \small
    \caption{Ablation study of feature extraction on CAS(ME)$^2$, comparing FEDN with I3D and variations of FEDN without attention modules.}
    \begin{tabular}{c|ccc|cc|c}
        \midrule[1.2pt]
        \multirow{2}{*}{} & \multicolumn{1}{c|}{\multirow{2}{*}{TP}} & \multicolumn{2}{c|}{\begin{tabular}[c]{@{}c@{}}F1\\ ($10^{-2}$)\end{tabular}} & \multicolumn{2}{c|}{\begin{tabular}[c]{@{}c@{}}mAP \\ ($10^{-3}$)\end{tabular}} & \multirow{2}{*}{\begin{tabular}[c]{@{}c@{}}\#training \\ params(M)\end{tabular}}                               \\ \cline{3-6}
                          & \multicolumn{1}{c|}{}                    & \textbf{\textit{S}}                                                                                      & \textbf{\textit{R}}                                                                                        & \textbf{\textit{S}}                                                              & \textbf{\textit{D}} &       \\ \midrule[1.2pt]
        I3D               & 122                                      & 45.9                                                                                                     & 69.7                                                                                                       & 88                                                                               & 35                  & 30.52 \\ \midrule[0.8pt]
        w/o seg. att.     & 116                                      & 47.6                                                                                                     & 75.9                                                                                                       & 99                                                                               & 47                  & 23.41 \\
        w/o sw att.       & 121                                      & 47.6                                                                                                     & 76.9                                                                                                       & 90                                                                               & 43                  & 24.99 \\
        w/o both          & 111                                      & 47.5                                                                                                     & 81.9                                                                                                       & 79                                                                               & 38                  & 23.41 \\
        w/o both(U2)      & 114                                      & 47.8                                                                                                     & 84.6                                                                                                       & 98                                                                               & 46                  & 31.81 \\
        FEDN(U2)          & 122                                      & 49.8                                                                                                     & 83.6                                                                                                       & 113                                                                              & 54.4                & 33.38 \\
        \textbf{FEDN}     & 129                                      & 51.1                                                                                                     & 75.2                                                                                                       & 106                                                                              & 51.6                & 24.98 \\ \midrule[1.2pt]
    \end{tabular}
    \label{ablation}
\end{table}

\subsection{Metric Selection of Spotting and Detection}
\label{sec:metric_selection}
While F1 score is widely used for spotting (e.g., in CAS(ME)$^2$), it only reflects performance at IoU 0.5 and a fixed confidence threshold, overlooking variability across other thresholds. In contrast, mAP offers a more comprehensive view by evaluating the full precision-recall curve over multiple IoUs. Recognition F1 (reported in \cite{liong2023spot}) can also be misleading, as lower spotting F1 may artificially inflate recognition scores due to fewer predicted intervals (see Table~\ref{ablation}). We thus advocate using mAP and IoU-based analyses for a more reliable assessment of both spotting and detection.

\subsection{Difference between CAS(ME)$^2$ and CAS(ME)$^3$}

Although CAS(ME)$^2$ and CAS(ME)$^3$ were collected by the same group, their dataset composition differs markedly. As shown in Table~\ref{performance}, model performance is significantly lower on CAS(ME)$^3$, likely due to the dominance of negative emotions such as fear and sadness, which often involve similar muscle movements.
This challenge is reflected in the confusion patterns: fear samples are frequently misclassified as sadness, and vice versa, with both categories showing high confusion with others like anger and disgust. For example, 96 fear samples were correctly predicted, but 97 were misclassified as sadness, nearly equal to the correct count. Such ambiguity contributes to a lower overall F1 score of 0.41 on CAS(ME)$^3$.
These results highlight the limitations of current models in distinguishing fine-grained emotions, the need for more robust annotation protocols, and the potential benefit of techniques like domain adaptation or multi-modal integration. The inconsistency between expert-annotated and self-reported labels further complicates dataset reliability.

\section{Conclusion}


We have introduced an end-to-end facial expression detection framework that uses shared attention-based representations to improve feature integration beyond cascaded pipelines. Experiments on CAS(ME)$^2$, CAS(ME)$^3$, and SAMMLong show state-of-the-art performance, with joint training consistently improving spotting accuracy over separate or two-stage approaches.

\vspace{1em}

\noindent \textbf{Acknowledgement} This work was supported in part by the Hong Kong Research Grants Council General Research Fund under grant number 16214821.

%
\bibliographystyle{IEEEtran}
\bibliography{citation}

\end{document}